\documentclass{svjour3}
\usepackage[hyphens]{url}
\usepackage{xspace}
\usepackage{color}
\usepackage{array}
\usepackage{underscore}
\usepackage{tipa}
\usepackage{cite}
\usepackage{stmaryrd}

\usepackage{rotating}

\usepackage{amsmath}
\usepackage{amssymb}
\usepackage{booktabs}
\usepackage{graphicx}
\usepackage{listings}
\usepackage{paralist}
\usepackage{subfig}
\usepackage{hyperref}
\usepackage{breakurl}
\usepackage{footnote}

\makeatletter

\def\holdocspecials{\do\ \do\$\do\&%
  \do\#\do\^\do\^^K\do\_\do\^^A\do\%}

\def\holtt{\trivlist \item[]\if@minipage\else\vskip\parskip\fi
\leftskip\@totalleftmargin\rightskip\z@
\parindent\z@\parfillskip\@flushglue\parskip\z@
\@tempswafalse \def\par{\if@tempswa\hbox{}\fi\@tempswatrue\@@par}
\obeylines \tt \let\do\@makeother \holdocspecials
 \frenchspacing\@vobeyspaces}

\makeatother

\newlength{\hsbw}
\newcommand\HOLSpacing{13pt}

   \newcommand\hilbert{\varepsilon}

   \newcommand{\Cond}{\(\rightarrow\)}
   \newcommand{\Eqv}{\(\equiv\)}
   \newcommand{\Iff}{\(\Longleftrightarrow\)\hspace{-1.5mm}}
   \newcommand{\Fa}{\(\forall\)}
   \newcommand{\Et}{\(\exists\)}
   \newcommand{\Eu}{\(\exists_{unique}\)}

   \newcommand{\Impl}{\(\Longrightarrow\)\hspace{-1.5mm}}
   \newcommand{\Func}{\(\to\)\hspace{-1.5mm}}

   \newcommand{\Lam}{\(\lambda\)}
   
   \newcommand{\Minus}{\(-\)}
   \newcommand{\Lminus}{\(-\)\hspace{-1.5mm}}
   \newcommand{\Prime}{\('\)}
   \newcommand{\Und}{\_}
   \newcommand{\Lt}{\(<\)}
   \newcommand{\Gt}{\(>\)}
   \newcommand{\Leq}{\(\leq\)}
   \newcommand{\Geq}{\(\geq\)}
   \newcommand{\Eq}{\(=\)}
   \newcommand{\Lrb}{\((\)}
   \newcommand{\Rrb}{\()\)}

   \newcommand{\Next}{\(\bigcirc\)}
   \newcommand{\Prev}{\(\ominus\)}
   \newcommand{\WPrev}{\(\widetilde{\bigcirc}\)}
   \newcommand{\Event}{\(\Diamond\)}
   \newcommand{\Once}{\(\underline{\Diamond}\)}  
\newcommand{\Hilbert}{\(\hilbert\)}

\newcommand{\Conj}{\(\wedge\)}
\newcommand{\Disj}{\(\vee\)}
\newcommand{\Neg}{\(\neg\)}
\newcommand{\Pnd}{\(\Diamond\)}

\newcommand{\Models}{\(\models\)}

\long\def\rechol#1#2#3{\let\next=\rechol\def\postnext{#2#3}\ifx#1\end
\let\next=\relax\def\postnext{\relax}
\else\ifx#1!\Fa                                          
\else\ifx#1@\Hilbert                                     
\else\ifx#1\#\Pnd                                        
\else\ifx#1'\Prime                                       
\else\ifx#1~\Neg                                         
\else\ifx#1\~\Neg
\else\ifx#1_\Und                                         
\else\ifx#1(\ifx#2+\ifx#3)\Next\def\postnext{}\fi        
            \else\ifx#2-\Prev\def\postnext{}             
            \else\ifx#2~\ifx#3)\WPrev\def\postnext{}\fi            
             \else\Lrb\fi\fi\fi                          
\else\ifx#1)\Rrb%
\else\ifx#1\/\Disj                                       
\else\ifx#1\.\Lam                                        
\else\ifx#1>\ifx#2=\Geq\def\postnext{#3}\else\Gt\fi      
\else\ifx#1?\ifx#2!\Eu\def\postnext{#3}\else\Et\fi       
\else\ifx#1-\ifx#2>\Func\def\postnext{#3}               
            \else\ifx#2-\Lminus\def\postnext{#3}
            \else\Minus\fi\fi                               
\else\ifx#1|\ifx#2-\Turns\def\postnext{#3}               
            \else\ifx#2=\Models\def\postnext{#3}
                 \else\Bar\fi\fi
\else\ifx#1<\ifx#2=\ifx#3>\Iff\def\postnext{}       
                   \else\Leq\def\postnext{#3}\fi    
            \else\ifx#2+\Event\def\postnext{}       
            \else\ifx#2-\Once\def\postnext{}       
            \else\Lt\fi\fi\fi                       
\else\ifx#1=\ifx#2=\ifx#3>\Impl\def\postnext{}            
                   \else\Eqv\def\postnext{#3}\fi         
            \else\ifx#2>\Cond\def\postnext{#3}
                 \else\Eq\fi\fi
\else\ifx#1/\ifx#2\^^M\Conj\par\def\postnext{#3}         
            \else\ifx#2\ \Conj\ \def\postnext{#3}\else#1\fi\fi  
\else#1\fi\fi\fi\fi\fi\fi\fi\fi\fi\fi\fi\fi\fi\fi\fi\fi\fi\fi\fi
\expandafter\next\postnext}

\newcolumntype{*}{>{\global\let\currentrowstyle\relax}}
\newcolumntype{^}{>{\currentrowstyle}}

\def\systemname#1{\textsf{#1}\xspace}

\def\th#1{\texttt{#1}\xspace}

\newcommand{\HOLLight}{\systemname{HOL Light}}

\newcommand{\Isabelle}{\systemname{Isabelle}}

\newcommand{\Mizar}{\systemname{Mizar}}
\newcommand{\MPTP}{\systemname{MPTP}}
\newcommand{\Vampire}{\systemname{Vampire}}
\newcommand{\Epar}{\systemname{Epar}}
\newcommand{\Z}{\systemname{Z3}}
\newcommand{\E}{\systemname{E}}
\newcommand{\V}{\systemname{V}}

\newcommand{\MML}{\systemname{MML}}
\newcommand{\MizAR}{\systemname{Miz$\mathbb{AR}$}}

\newcommand{\Flyspeck}{\systemname{Flyspeck}}

\newcommand{\BliStr}{\systemname{BliStr}}
\newcommand{\gensim}{\systemname{gensim}}

\newcommand{\SInE}{\systemname{SInE}}
\newcommand{\MaLARea}{\systemname{MaLARea}}

\newcommand{\OCaml}{\systemname{OCaml}}

\newcommand{\HH}{\systemname{HOL(y)Hammer}}

\title{MizAR 40 for Mizar 40}
\author{Cezary Kaliszyk \and Josef Urban}

\institute{Cezary Kaliszyk \at University of Innsbruck, Austria \and
  Josef Urban %
\at  Radboud University, Nijmegen}

\begin{document}
\maketitle
\begin{abstract}
  As a present to \Mizar on its 40th anniversary, we develop an AI/ATP
  system that in 30 seconds of real time on a 14-CPU machine
  automatically proves 40\% of the theorems in the latest official
  version of the Mizar Mathematical Library (\MML). This is a
  considerable improvement over previous performance of large-theory
  AI/ATP methods measured on the whole \MML. To achieve that, a large
  suite of AI/ATP methods is employed and further developed.  We
  implement the most useful methods efficiently, to scale them to the
  150000 formulas in \MML. This reduces the training times over the corpus to 1-3 seconds,
  allowing a simple practical deployment of the methods in the 
  online automated reasoning service for the \Mizar users (\MizAR).

\end{abstract}

\section{Introduction and Motivation}
\label{introduction}

Since 2003 the \Mizar Mathematical Library\footnote{\url{http://www.mizar.org}} (\MML) has been used as a
repository for developing AI/ATP methods for solving formally
stated (computer-understandable) conjectures in general large-theory mathematics~\cite{Urban06,Urb04-MPTP0,Urban03}.
The number and strength of the methods developed has been growing,
however the methods were often developed and evaluated on smaller
benchmarks such as the MPTP Challenge\footnote{\url{http://www.cs.miami.edu/~tptp/MPTPChallenge/}} and MPTP2078\cite{abs-1108-3446}. 
Recently, we have tried to develop a strong suite of AI/ATP methods that scale to the
whole June 2012 version of the \Flyspeck~\cite{Hales05}
development~\cite{holyhammer,EasyChair:74}, containing more than 14000
theorems. The best methods using the accumulated data have been
recently deployed in an online (``cloud-based'') AI/ATP service for
\HOLLight~\cite{Harrison96} formalizations~\cite{KaliszykU13,hhmcs}. When running the 14
strongest methods in parallel, 47\% of the \Flyspeck theorems can be
proved in 30 seconds without any user interaction. To a significant extent,
this performance is achieved by learning from the large number of previous proofs 
various high-level~\cite{Urban11-ate,KuhlweinLTUH12} and low-level~\cite{blistr,EasyChair:74} guiding 
methods for state-of-the-art ATP and SMT systems such as \Vampire~\cite{Vampire}, \E~\cite{Sch02-AICOMM} and \Z~\cite{z3}.
A similar work has been recently undertaken with \Isabelle~\cite{KuhlweinBKU13}.

We believe that this performance is a milestone on the way to John
McCarthy's AI and QED dream of ``Heavy Duty Set Theory'', i.e., a sufficiently smart
AI/ATP/ITP system that can without forcing mathematicians to struggle
with various current technologies of explicit ``proof programming''
automatically understand and check reasoning steps done on the level
that is commonly used in mathematical proofs. If such a system is
developed, the current barrier preventing computer understanding of
common mathematical proofs will to a large extent disappear, and
mathematics (and thus all human exact thinking) may enter an era of
ubiquitous computer understanding and strong AI assistance.

In this work we employ, further develop, and evaluate a suite of
scalable AI/ATP methods on the whole \Mizar library, containing nearly
58000 theorems. The main experimental result (Section~\ref{Results})
is that the 14 strongest methods run in parallel for 30 seconds prove
40.6\% of the 58000 \Mizar theorems without any user interaction. If
users are also allowed to manually select the relevant premises used
then by the ATPs, the performance grows to 56.2\%. Our hope is that this
performance may significantly lower the barrier to formalizing
mathematics in the \Mizar system~\cite{mizar-in-a-nutshell}, which has a long history of
targeting mathematicians with its standard logical foundations,
intuitive proof style, and linguistic closeness to mathematical
vernacular. The various methods used to achieve this performance
are described in Section~\ref{Methods}. Section~\ref{Results} discusses the experiments and results obtained with the methods, 
Section~\ref{Proof} takes a
brief look at the data obtained from the automatically found proofs,
Section~\ref{Integration} 
briefly describes the first integration of the methods in the \MizAR online service~\cite{abs-1109-0616},
and Section~\ref{Future} discusses future work and concludes.

\section{Learning Proof Guidance From the \MML}
\label{Methods}
The general idea behind the large-theory ATP-for-ITP systems that
started to be developed in the last decade is to combine (i)
translations between the ATP and ITP formalisms with (ii) high-level
knowledge selection methods and (iii) state-of-the-art ATP systems
which can be further strengthened and tuned in various ways for the
large-theory setting.  For the translation from the \Mizar logic to
TPTP we re-use the existing MPTP translation~\cite{Urban06,Urb04-MPTP0}. After several initial
experiments with various ATPs and their versions, we have decided to limit the set of
ATPs to \Vampire 3.0, \Z 4.0, and \E 1.8 run using the \Epar scheduler
and strategies~\cite{blistr}. This combination also worked well for the experiments with \Flyspeck. 
In this work, the main focus is on (ii),
i.e., on deploying and improving for \MML the suite of scalable high-level
knowledge selection methods which we have recently developed for the
whole \Flyspeck corpus, containing over 20000 formulas.

The currently strongest knowledge selection methods for large-theory
mathematics are \emph{data-driven}~\cite{KMB}: instead of explicit
programming of all aspects of knowledge selection, data-driven methods
extract (learn) significant parts of such complicated algorithms from
the existing large libraries of solutions (proofs).  This shift from
attempts at explicit (\emph{theory-driven}) programming of AI
heuristics to learning AI algorithms from data is to a large extent
responsible for the recent successes in hard AI domains such as web
search, consumer choice prediction, autonomous car driving, etc. But
this also means that extracting good training data from the \MML is
equally important as the methods that learn knowledge selection on
such data. Interestingly, in large-theory ATP there is a full positive
feedback loop~\cite{Urban07} between the amount/quality of the data and the strength
of the methods: not only more/better data produce stronger methods
(which is the standard data-driven argument), but also stronger
proving methods produce more/better data in the form of proofs. This
is quite a unique property of this very expressive and fully semantic
AI domain, born quite recently thanks to the development on large
formal mathematical libraries such as the \MML~\cite{UrbanV13}. The
main body of our work thus consists of the following
steps:

\begin{enumerate}
\item Obtaining from the \MML suitable data (proof dependencies) on
  which premise selection methods can be trained.
\item Developing, training and testing such premise selection methods
  and their parameters on a small random subset of the \MML.
\item Iterating steps (1) and (2), i.e., using the most successful
  methods to get more proofs, and training further methods on them.
\end{enumerate}

\subsection{Obtaining Proof Dependencies from the \MML}
\label{Deps}

There are 57897 \Mizar theorems and unnamed toplevel lemmas in the
most recent official \MML 4.181.1147. This set is canonically ordered
by the \MML order of articles, and by the order of theorems in the
articles. This ordering also applies to the about 90.000 other \Mizar
formulas used in the problems. Our goal is to prove automatically as
many of the 57897 theorems as possible, using at each point all the
available formulas and all information about previous theorems and
their proofs.

The human-written \Mizar proofs contain explicit information about the
theorems and definitions used. This information is however
incomplete. For re-playing the \Mizar proofs with ATPs, a lot of
``background'' knowledge (typically about typing of terms) needs to be
explicitly added. The \MPTP system adds such background formulas
heuristically in a fixpoint manner, by looking at the set of symbols
in the problem and adding the appropriate typing formulas. The average
size of an ATP problem constructed in this way by \MPTP is 328
formulas, while the average number of the explicit \Mizar proof
references is only 12. In~\cite{abs-1108-3446} we have constructed a
computationally expensive method (using the \Mizar checker) that
reduces the number of the background formulas 2-3 times. However, the
measured performance gain from that method when re-proving with ATPs
the MPTP2078 problems was only about 4\%, and in~\cite{AlamaKU12} it
was found that the ATPs still typically do not use many of the
\Mizar-needed background formulas. Since learning from the minimized ATP proofs
is typically superior~\cite{KuhlweinLTUH12,holyhammer}, in the current
work we decided to skip the expensive \Mizar-based proof minimization, 
and focus on using the ATP proofs of the heuristically constructed re-proving problems coming from the \Mizar
theorems. This decision was influenced by older preliminary 20-second
testing using an Intel Xeon 2.67 GHz server, in which \Vampire 1.8 proved 20302
of the \MPTP-constructed theorem problems, \Epar proved 20324, and together they proved 23141,
i.e., 40\% of the theorems.

Table~\ref{Data1} shows the gradual growth of the set of ATP-computed
proof dependencies that we mainly use for learning. The first set is obtained
by running \Vampire 3.0 for 300 seconds on the \MPTP-constructed
re-proving problems. The additional 1677 solutions in the second set
are obtained by learning premise selection on the first set, and
running ATPs for 120 seconds again on various most relevant slices of
the re-proving problems (we always include the explicit \Mizar references in
such pruned ATP problems). The following passes no longer prune the
\MPTP-constructed re-proving problems. They just use the premise
selectors trained on the previous passes to suggest the most relevant
premises, regardless of the original \Mizar proofs.  These
\MaLARea-style~\cite{Urban07,US+08} iterations are also quite expensive to compute for a new
development. However, we have shown in~\cite{hhmcs} that this
information can be in large libraries efficiently re-used and does not
need to be computed for every version again. As in~\cite{holyhammer},
the difference to the original \MaLARea iterations is that at each
premise-selection point only the chronologically previous proofs are
used for learning. This corresponds to the ultimate deployment
scenario, when always only the library proofs written so far are
known.

\begin{table}[bhtp]
\caption{Improving the dependency data used for training premise selection}
\centering
\begin{tabular}{lcccc}\toprule
Pass & ATPs & Premise Selection & Theorems (\%)& Dependencies \\\midrule
1 & \V  300s & \MPTP re-proving &     27842 (48\%)& 27842 \\
2 & \V, \E, \Z 120s & trained on (1), premises limited by (1) & 29519 (51\%)& 30024 \\ 
3 & \V, \E 120s & trained on (2), premises unlimited & 30889 (53.4\%)  & 31720 \\
4 & \V, \E 120s & trained on (3), premises unlimited &  31599 (54.6\%)& 32976 \\
5 & \V, \E 120s &  trained on (4), premises unlimited &  32010 (55.3\%)& 35870 \\
6 & \V, \E 120s &  trained on (5), premises unlimited &  32165 (55.6\%) & 36122 \\\bottomrule
\end{tabular}
\label{Data1}
\end{table}

In total, these iterations yield 32165 ATP proofs, and with the final
evaluation described in Section~\ref{WholeMML} this number reaches
32557 theorems. This means that when using either human or AI-based
premise selection and their combinations, state-of-the-art ATPs are
today able to prove 56.23\% of the toplevel \MML theorems. This is a very good
motivation for developing good premise-selection methods.

\subsection{Premise Selection Techniques}
The premise selection techniques we start with are the relatively fast
scalable methods used for \Flyspeck in~\cite{holyhammer}: Naive Bayes
(\th{nb}) and distance-weighted k-nearest neighbor~\cite{DudaniS76}
(\th{knn}). In particular, a family of differently parametrized k-NNs
together with the IDF (inverse document frequency) feature weighting
scheme~\cite{Jones72astatistical} have recently provided quite
significant performance improvement~\cite{EasyChair:74}.  This is here
extended to naive Bayes (\th{nb_idf}).  We are interested both in the
strongest possible methods, and also in methods that can be quite
weak, but complement well the stronger methods.

Apart from minor implementational modifications, we characterize each
formula with the syntactic features used by \MaLARea: symbols, terms
and subterms of the formula. In the most successful methods, all
variables in such features are renamed to just one variable \th{A0}
(widening the similarity relation), however to a smaller extent, also
the features with original variables are useful. Following the recent
successful use by \MaLARea 0.5 in the 2013 CASC LTB
competition\footnote{\url{http://www.cs.miami.edu/~tptp/CASC/24/WWWFiles/DivisionSummary1.html}},
we also add a version of distance-weighted k-NN using the Latent
Semantic Indexing~\cite{DeerwesterDLFH90} (LSI) preprocessing of the
feature space done efficiently by the \gensim~\cite{rehurek10}
toolkit. We test the LSI preprocessing with 800, 3200 and 6400 topics (\th{lsi_800} .. \th{lsi_6400}),
and also versions with and without the TF-IDF feature scaling (e.g., (\th{lsi_3200ti})).

The next modification of k-NN are various recursive schemes for
weighting the neighbors' dependencies. The \th{geo_1_F} version stops
the dependency recursion at the first level, weighting each dependence
of a neighbor $N$ by the factor $F*distance(N)$ (where $F \in (0,1)$),
and taking maximum (instead of sum) of such weights over all
neighbors.  The \th{geo_r_F} version does full dependency recursion,
weighting the indirect dependencies by
$F^{recursionlevel}*distance(N)$, and again taking maximum over all
such factors.

In~\cite{KuhlweinLTUH12} a linear combination of the strongest
learning method with the \SInE~\cite{HoderV11} heuristic produced very
good results. This is an instance of \textit{ensemble learning} where
multiple base methods are combined into stronger classifiers. We
heuristically explore combinations of the various base methods using
various weighting schemes. In addition to the linear combination, we
try geometric, harmonic, and quadratic average, and also use minimum
and maximum of ranks. In particular taking the minimal rank
(\th{comb_min}) and the geometric average (\th{comb_geo}) of ranks
(computed as additions of logarithms) turn out to be quite
successful. This can be explained in various ways, for example, taking
the geometric average is the correct way of averaging ratios. Since
ATPs are very (probably exponentially) sensitive to the number of
axioms, treating the particular aggregated rankings as ratios is quite
likely fitting to our domain (e.g., the ratio between 50th
and 60th premise is 1.2, while the ratio between 10th and 20th
premise is 2, whereas the linear distance is the same in the two cases).

We also try several methods of \textit{boosting}~\cite{Schapire90}: using for training
of the next method only those proof dependencies that are badly predicted by the
previous methods. While we believe that there are good reasons why
this approach should help (e.g., our current methods being quite simple
and thus hard to fit to more complicated ideas), so far this has not
provided significant improvements.

All the tested methods\footnote{For their details see \url{http://cl-informatik.uibk.ac.at/users/cek/mizAR/legend.txt}} (apart from LSI (\gensim) and \SInE (\E
prover)) are now uniformly implemented in \OCaml, which gives significant
speedup over the initial Perl implementation on the large number of
features, labels and examples used when training over the whole \MML
(the number of features reaches several hundred thousand). The most
useful methods are further implemented in C++, making them about twice
as fast as their \OCaml version. A particularly useful low-level
optimization is the use of partial sorting (based on heapsort) of the
scores according to the number of premises demanded from a particular
premise-selection method. For example, if only 128 premises are
needed, the partial sorting is much more efficient than full sorting
of the whole array of 150000 \MML formulas.

\subsection{ATPs and Their Low-Level Guidance}

No particular development of ATP strategies was done for this
work on the whole \MML. However, thanks to the recent CASC competitions containing
\Mizar divisions (Mizar@Turing12, CASC LTB 2013) the recent versions
of \Vampire and \Epar seem to be tuned well for MPTP2078 . In
particular, a set of strong strategies for \E has been automatically
developed by \BliStr~\cite{blistr} in 2012 on the 1000 Mizar@Turing12
problems, raising the performance on MPTP2078 over \E's auto-mode by
25\%. A second round of such strategy evolution on these 1000
problems was done for \MaLARea 0.5 in CASC LTB 2013, where
additionally a number of strong \SInE strategies were evolved. 
\Vampire 3.0 is on \MML 16\% stronger than \Vampire 2.6.

\section{Experiments and Results}
\label{Results}

\subsection{Experiments and Results on a Random Subset of 1930 Problems}
Most of the experimental research was done on a
random subset of \MML consisting of 1930 theorems (more precisely, every 30th theorem was used). For each of these
theorems, the premise selection methods were trained on all the
preceding proof data (Section~\ref{Deps}), and a chosen numbers (32,
64, 96, 128, 256, 512 and 1024) of the best-ranked premises were given
to the ATPs. Most of the experiments during the development of the
premise selection methods were done with \Vampire 3.0. The final 
experiments were extended to \Epar and \Z.  As
in~\cite{holyhammer}, the systems were run with a 30 second time limit
on a 48-core server with AMD Opteron 6174 2.2 GHz CPUs, 320 GB RAM,
and 0.5 MB L2 cache per CPU.  In real time, each evaluation thus took
2-3 hours for one ATP. In total, 70 different premise-selection methods
have been tried on the random sample, see our web page for a detailed
listing.\footnote{\url{http://cl-informatik.uibk.ac.at/users/cek/mizAR/5yp.html}}

Table~\ref{Best1930} shows the fourteen methods and their
parameters that performed best on the 1930-subset.  The
$\mathrm{\Sigma}$-SOTAC (State of the art contribution) is the sum of
a system's SOTAC over all problems attempted, where for each problem
solved by a system, its SOTAC for the problem is the inverse of the
number of systems that solved the problem. This metric shows how
useful a particular method is in a collection of other methods (in
this case all the 70 methods that have been tried).

\begin{table}[bhtp]
\caption{14 best premise selection methods on the 1930-subset}
\centering
\begin{tabular}{>{\ttfamily}lcccccc}\toprule
Method & Parameters&  Premises & ATP	& Theorem (\%) & $\mathrm{\Sigma}$-SOTAC & Processed \\\midrule
comb&	min_2k_20_20	 &	 128 &\Epar&550	(28.50)&	2.41& 1930\\ 
comb&	geo_3k_50_00	 &       96&	\V&544	(28.19)&	1.94& 1930\\ 
lsi &    3200ti_8_80	 &       128&\Epar&537	(27.82)&	2.17& 1930\\ 
comb&	geo_1k_33_33	 &       96&	\V&535	(27.72)&	1.60& 1930\\ 
comb&	geo_10k_33_33	 &       96&	\V&533	(27.62)&	1.73& 1930\\ 
comb&	geo_3k_20_20	 &       96&	\V&533	(27.62)&	1.71& 1930\\ 
comb&	geo_3k_33_33	 &       96&	\V&533	(27.62)&	1.63& 1930\\ 
comb&	geo_3k_25_25_25	 &       96&	\V&532	(27.57)&	1.62& 1930\\ 
comb&	har_2k_k200_33_33&	 256 &\Epar&532	(27.57)&	2.19& 1930\\ 
comb&	geo_3k_33_33	 &       96&	\V&531	(27.51)&	1.69& 1930\\ 
comb&	geo_3k_50_00	 &       128&	\V&531	(27.51)&	1.76& 1930\\ 
comb&	geo_1k_33_33	 &       128&	\V&529	(27.41)&	1.68& 1930\\
knn &    is040		 &       128&	\V&528	(27.36)&	1.85& 1930\\
knn &    is_40		 &       96&\Epar&528	(27.36)&	2.40& 1930\\\bottomrule
\end{tabular}
{\small
  \begin{description}
  \item[Theorem (\%):] Number and percentage of theorems proved by a system.
  \item[$\mathrm{\Sigma}$-SOTAC:] See the explanatory text for this metric.
  \end{description}}
\label{Best1930}
\end{table}

Table~\ref{Greedy1930} shows the 14 methods that collectively (when
computed in a greedy way) cover as many of the 1930 problems as
possible. To be as orthogonal as possible, the methods in this set
differ a lot in their parameters, the data trained on, and the number
of best premises given to the ATP. Their joint performance on this
subset is 44\%. All the 70 tested methods together solve 968 of the
1930 problems, i.e., 50.155\% .

\begin{table}[bhtp]
\caption{The top 14 methods in the greedy sequence on the 1930-subset}
\centering
\begin{tabular}{>{\ttfamily}lcccccc}\toprule
Method & Parameters& Deps &  Premises & ATP	 & Sum \% & Sum \\\midrule
comb&min_2k_20_20&ATP6 &0128&\Epar	&28.497	&550	      \\           
comb&qua_k200_3k_33_33&ATP4 &0512 &\V&32.798	&633	      \\           
comb&geo_3k_33_33&ATP3 &0064&\V	&35.959	&694	      \\           
lsi &3200ti_8_80&ATP6  &0128&\Z &37.461	&723	      \\           
geo &r_99&ATP6        &0064&\V	        &38.653	&746  \\             
knn &200 &ATP1+\Mizar          &1024&\V	        &39.741	&767  \\             
nb  &idf010&ATP6      &0128&\Epar	        &40.725	&786  \\             
comb&min_20_20&ATP2   &0128&\V	&41.347	&798	      \\           
comb&geo_3k_50_00&ATP3&1024&\V	&41.969	&810	      \\           
knn &is040&ATP1        &1024&\Epar	        &42.487	&820  \\             
knn &is_40&ATP6       &0096&\Z	        &43.005	&830  \\             
geo &1_66&ATP3        &1024&\V	        &43.420	&838  \\             
lsi &6400_8_120&ATP1   &0064&\V	        &43.782	&845  \\             
geo &1_33&ATP3        &0256&\V	        &44.093	&851  \\\bottomrule  
\end{tabular}
\label{Greedy1930}
\end{table}

\subsection{Experiments and Results on the Whole \MML}
\label{WholeMML}
When the developed methods on the 1930-subset reached sufficiently
high joint performance, we evaluated the most useful 14 methods on the
whole \MML, again with a 30-second time limit. This took about one
week of real time on our server. The performance of these methods is
shown in Table~\ref{PerfOnMML}. The methods are ordered there from top
to bottom already by their position in the greedy covering sequence
for the whole \MML. The table says that running these fourteen
methods in parallel for 30 seconds gives a 40.6\% chance of solving an
\MML theorem without any user interaction. 
The best previous result in such fully automated learning/proving over
the whole \MML was 18\%, achieved in~\cite{AlamaKU12} by running
\Vampire 0.6 for 20 seconds (using about twice as fast Intel Xeon
machine than our AMD server) on 200 best-ranked premises proposed by
the SNoW system using the Naive Bayes learner. Since this was just a
single method, a fair comparison is with the best method developed
here, which solves 27.3\%, i.e. 50\% more problems. One of the reasons for this 
improvement are obviously the better training data developed here by the 
six \MaLARea-style proving/learning passes over the \MML.

It should be however noted that much better results than 18\% have
been achieved on smaller benchmarks such as MPTP2078, where more
expensive methods such as kernel-based learning~\cite{KuhlweinLTUH12}
could be applied. Comparison with those results is however possible
only in a high-level way: we use different \MML version here,
different versions of the ATPs, courser slices of the best premises,
and we do not limit the premises only to those available in the 33
articles used for MPTP2078. The best result on MPTP2078 reported
in~\cite{KuhlweinU12b} was 823 problems (out of 2078) solved with 70
premises, \Vampire 0.6 and 5s on an Intel Xeon machine. The best new
performance on the 2061 problems corresponding to the 33 MPTP2078
articles in the current \MML is 1059 problems solved in 30 seconds by
\Epar using 128 best premises.\footnote{The detailed results
  restricted to the 2061 problems are at
  \url{http://cl-informatik.uibk.ac.at/users/cek/mizl/mptp2k.html}.}
To make a bit closer comparison, we test the current best-performing
method on the 2061 problems also with the old \Vampire 0.6 and 5
seconds on the old Intel machine, solving 726 problems. This is
practically the same result as the performance of the best old
kernel-based method (combined with SInE) on the MPTP2078 benchmark
when using 128 premises. This seems to be an evidence (modulo all the
differences named above) that the methods based on fast scalable
learning techniques such as k-NN can with enough care catch up with
the existing kernel-based techniques. Quite likely, this is however
not the last word, and we hope to get further improvements by scaling
up and strengthening the kernel-based and related methods.

\begin{table}[bhtp]
\caption{14 most covering methods on the whole \MML, ordered by greedy coverage.}
\centering
\begin{tabular}{>{\ttfamily}lcccccc}\toprule
Method & Parameters&  Prems. & ATP	& $\mathrm{\Sigma}$-SOTAC & Theorem (\%) & Greedy (\%)\\\midrule
comb&	min_2k_20_20	&0128	&\Epar	&1728.34	&15789	(27.3)&15789  (27.2)	      \\           
lsi &    3200ti_8_80	&0128	&\Epar	&1753.56	&15561	(26.9)&17985  (31.0)	      \\                      	
comb& qua_2k_k200_33_33	&0512	&\Epar	&1520.73	&13907	(24.0)&19323  (33.4)	      \\                      	
knn &      is_40	&0096	&\Z	&1634.50	&11650	(20.1)&20388  (35.2)	      \\                          
nb  &     idf010	&0128	&\Epar	&1630.77	&14004	(24.2)&21057  (36.4)	      \\                      	
knn &      is_80	&1024	&\V	&1324.39	&12277	(21.2)&21561  (37.2)	      \\                      	
geo &       r_99	&0064	&\V	&1357.58	&11578	(20.0)&22006  (38.0)	      \\
comb&	geo_2k_50_50	&0064	&\Epar	&1724.43	&14335	(24.8)&22359  (38.6)	      \\                      	
comb&	geo_2k_60_20	&1024	&\V	&1361.81	&12382	(21.4)&22652  (39.1)	      \\                      	
comb& har_2k_k200_33_33	&0256	&\Epar	&1714.06	&15410	(26.6)&22910  (39.6)	      \\                      	
geo &       r_90	&0256	&\V	&1445.18	&13850	(23.9)&23107  (39.9)	      \\                      	
lsi &    3200ti_8_80	&0128	&\V	&1621.11	&14783	(25.5)&23259  (40.2)	      \\                      	
comb&	geo_2k_50_00	&0096	&\V	&1697.10	&15139	(26.1)&23393  (40.4)	      \\                      	
geo &       r_90	&0256	&\Epar	&1415.48	&14093	(24.3)&23478  (40.6)	      \\\bottomrule                      	
\end{tabular}
\label{PerfOnMML}
\end{table}

\section{Proofs}
\label{Proof}
We have briefly compared the shortest ATP proofs found with the
corresponding \MML proofs. For this, we only consider the 28892 named
\Mizar theorems for which we have obtained either a human or
AI-advised ATP proof. The complexity metric used for a human-written
\Mizar proof is just the number of proof lines in the \Mizar article,
while for the ATP proofs we use the number of
dependencies.\footnote{These choices can obviously be questioned, but
  as a first comparison they are useful enough.}  This statistics,
sorted by the largest difference between these metrics is available
online\footnote{\url{http://mizar.cs.ualberta.ca/~mptp/mml4.181.1147/html/00prdiff15.html}},
together with the ATP dependencies used for this
comparison\footnote{\url{http://mizar.cs.ualberta.ca/~mptp/mml4.181.1147/html/00atpdeps}}.
For example the first entry says that the ATP proof of the theorem
\th{REARRAN1:24}\footnote{\url{http://mizar.cs.ualberta.ca/~mptp/mml4.181.1147/html/rearran1.html\#T24}}
has a 534-lines long \Mizar proof, while the shortest ATP proof found
has only 5 dependencies. Indeed, this greatest AI/ATP-found proof
shortening is valid, thanks to a symmetry between the concepts used in
this theorem and a previously proved theorem
\th{REARRAN1:17}\footnote{\url{http://mizar.cs.ualberta.ca/~mptp/mml4.181.1147/html/rearran1.html\#T17}}
which can be established quite quickly from the concepts'
definitions. The \Mizar proof instead proceeds by repeating the whole
argument from scratch, modifying it at appropriate places to the
symmetric concepts. The AI/ATP toolchain has thus managed to
succinctly express the difference between the two theorems in a very
explicit and operational way, while the human authors probably were on
some level also aware of the symmetry, but were not able to capture it
so precisely and succinctly. In some sense, the AI/ATP system has thus
managed to find, precisely formulate, and productively use a new
mathematical trick.
 
This comparison, showing such most striking
shortenings, is also useful for heuristic checking of the correctness of
the whole translation/AI/ATP toolchain. By random inspection of a
dozen of such shortenings, no suspicious proofs were found, i.e., all
the inspected ATP proofs could be replayed in \Mizar. On the other
hand, some of the ATP proofs can get very long, and may be probably
already quite hard to understand without further refactoring and
presentation methods.

Finally, the \Mizar proof length expressed in terms of the lines of
code can also serve as another metric for measuring the performance of
the ATP methods. The total number of the \Mizar source lines used for
the proofs of the 52248 named toplevel theorems is 1297926. The sum of
the \Mizar proof lines of the 28892 named theorems that were proved
automatically (either from human or AI-selected premises) is
300914. This means that on average 23.2\% of the proof lines can be
``written automatically'', if such automation is called on the
toplevel named theorems. This is a metric that in some sense
complements the 56.23\% ratio obtained in Section~\ref{Deps}, showing
that the ATPs are much better in proving the \Mizar-easy theorems.  On
the other hand, the 23.2\% average would be obviously improved a lot
if also the proof-local lemmas were included in the experiments, and
the number of lines corresponding to such lemmas was appropriately
included in the statistics.

\section{Integration with \MizAR}
\label{Integration}

The new optimized C++ versions of the premise selectors are sufficiently fast to
train on the whole \MML in 1-3 seconds. This simplifies the integration of the methods in the \MizAR service. For each query, the premise
selectors are always first trained on the whole \MML and also on the features
and proof dependencies added from the current article.\footnote{In the \MizAR service, the conjecture is always submitted with the whole \Mizar article in which the conjecture is stated.} After such
training, the premise selectors are presented with the conjecture
features, to which they respond by ranking the available theorems
according to their relevance for the conjecture. The several premise selection
methods with their corresponding ATPs are run in parallel, and if successful, the
result is communicated to the user. The main \MizAR server (Intel Xeon
2.67 GHz) is considerably faster than the AMD machines used for most
of the experiments. 

The service thus always updates itself with new data: the conjecture
is always a part of a particular \Mizar article, which is submitted as
a whole to the system. However, in comparison with the recently
produced \HOLLight service (\HH), the updating is so far more
limited. We do not yet try to get (for better training) the minimized
ATP proofs of the article's theorems that precede the current
conjecture. One reason is that, unlike in \HH, the \MizAR service
allows anonymous uploads of whole articles, but does not yet keep such
projects persistent. Adding such persistence should make the computing
and minimization of ATP proofs less expensive, because such data can
then be quite efficiently cached and re-used (see Section 3 of~\cite{hhmcs}). 

Another difference to the \HOLLight setting is the very common use of
local constants (eigenvariables) in the Ja\'skowski-style \Mizar
proofs. The large-scale experiments (and thus also the training data
obtained from them) presented here only deal with the set of toplevel
\Mizar theorems which do not contain such proof-local constants. This
has two different effects when proving the proof-local lemmas that contain such
constants: 
\begin{enumerate}
\item The local assumptions and lemmas about such constants are
  naturally preferred by the premise selectors (in particular when
  using weighting schemes such as TF-IDF~\cite{EasyChair:74}), because
  such local constants (which always have a distinct internal name)
  and the terms containing them are rare. This is good, because such
  local lemmas are typically quite relevant to the local conjecture.
\item The feature representation of the proof-local lemmas may become
  quite distant (in the various metrics used by k-NN) from the general
  theorems that are needed to justify such lemmas, because many terms
  in the lemmas are instantiated with the local constants. This may be
  a serious problem, preventing finding the relevant theorems.
\end{enumerate}
We use a simple
method to counter (2): In addition to the standard term features, we
also generalize all local constants in such proof-local conjectures to
variables, and use also all the term features of such generalized version of the lemmas.
This is clearly just a first step, the general task of getting
features that indicate for example the (lack of) the instantiation
relationship between two formulas is quite interesting, and various
syntactic and semantic methods are possible.  Further experiments and
evaluation of such issues, as well as of the user-perceived
strengthening of the \MizAR service are left as future work.

\section{Conclusion, Future Work and Thanks}
\label{Future}

The main result of this work is the 40.6\% success rate in proving the
toplevel \Mizar theorems fully automatically. This has been achieved
by several iterations of implementing better premise selection
methods, using them to obtain better training data, and using such
data to further improve the performance of the learning methods. The
methods were implemented very efficiently, allowing their easy
deployment in the \MizAR service. We believe that such strong AI/ATP
systems are very useful tools that make formal mathematics much more
accessible, and their gradual strengthening is today one of the most
promising paths towards the eventual adoption of computer-assisted
mathematics (and exact science) by mainstream mathematicians (and
exact scientists).

The main body of future work is thus further strengthening of the
various parts (e.g., features/labels and the whole learning setup,
machine-learning techniques, ATPs) of the AI/ATP methods.  Also, more advanced
proof reconstruction such as~\cite{proch,EasyChair:111} is still
missing for \Mizar. With longer and longer ATP proofs, human-friendly
transformations and presentations of such proofs are becoming more and
more important tasks that will quite likely also benefit from learning
the ``human-friendliness'' from large repositories of
human-oriented proofs such as the \MML. 

Thus it seems that the forty years of incessant and stubborn designing and
building of the human-oriented formal mathematical language and large
library by the \Mizar team, and in particular by the recently deceased
\Mizar gurus Andrzej Trybulec and Piotr Rudnicki, have already
resulted in one of the most interesting AI corpora currently available
to the mankind.  It will be quite hard for the historians to properly
enumerate all their inventions that led to the current state of the
art. We would like to thank Andrzej and Piotr for this lifelong Opus
Magnum, for their infatuating dreams, their wide and never-ending
interest in science (and science fiction), and for their great sense
of fun combined with high doses of self-criticism, down-to-earth
common sense, caution and modesty, that made them into such great
scientists, hackers, teachers, debaters, critics, and friends.

\bibliography{ate11}

\begin{thebibliography}{10}

\bibitem{abs-1108-3446}
Jesse Alama, Tom Heskes, Daniel K\"{u}hlwein, Evgeni Tsivtsivadze, and Josef
  Urban.
\newblock Premise selection for mathematics by corpus analysis and kernel
  methods.
\newblock {\em Journal of Automated Reasoning}, 2013.
\newblock \url{http://dx.doi.org/10.1007/s10817-013-9286-5}.

\bibitem{AlamaKU12}
Jesse Alama, Daniel K{\"u}hlwein, and Josef Urban.
\newblock {Automated and Human Proofs in General Mathematics: An Initial
  Comparison}.
\newblock In Nikolaj Bj{\o}rner and Andrei Voronkov, editors, {\em LPAR},
  volume 7180 of {\em LNCS}, pages 37--45. Springer, 2012.

\bibitem{z3}
Leonardo~Mendon\c{c}a de~Moura and Nikolaj Bj{\o}rner.
\newblock {Z3: An Efficient SMT Solver}.
\newblock In C.~R. Ramakrishnan and Jakob Rehof, editors, {\em TACAS}, volume
  4963 of {\em LNCS}, pages 337--340. Springer, 2008.

\bibitem{DeerwesterDLFH90}
Scott~C. Deerwester, Susan~T. Dumais, Thomas~K. Landauer, George~W. Furnas, and
  Richard~A. Harshman.
\newblock {Indexing by Latent Semantic Analysis}.
\newblock {\em JASIS}, 41(6):391--407, 1990.

\bibitem{DudaniS76}
Sahibsingh~A. Dudani.
\newblock The distance-weighted k-nearest-neighbor rule.
\newblock {\em Systems, Man and Cybernetics, IEEE Transactions on},
  SMC-6(4):325--327, 1976.

\bibitem{mizar-in-a-nutshell}
Adam Grabowski, Artur Korni{\l}owicz, and Adam Naumowicz.
\newblock {M}izar in a nutshell.
\newblock {\em Journal of Formalized Reasoning}, 3(2):153--245, 2010.

\bibitem{Hales05}
Thomas~C. Hales.
\newblock Introduction to the {F}lyspeck project.
\newblock In Thierry Coquand, Henri Lombardi, and Marie-Fran\c{c}oise Roy,
  editors, {\em Mathematics, Algorithms, Proofs}, volume 05021 of {\em Dagstuhl
  Seminar Proceedings}. Internationales Begegnungs- und Forschungszentrum
  f{\"u}r Informatik (IBFI), Schloss Dagstuhl, Germany, 2005.

\bibitem{Harrison96}
John Harrison.
\newblock {HOL Light}: A tutorial introduction.
\newblock In Mandayam~K. Srivas and Albert~John Camilleri, editors, {\em
  FMCAD}, volume 1166 of {\em LNCS}, pages 265--269. Springer, 1996.

\bibitem{HoderV11}
Krystof Hoder and Andrei Voronkov.
\newblock Sine qua non for large theory reasoning.
\newblock In Nikolaj Bj{\o}rner and Viorica Sofronie-Stokkermans, editors, {\em
  CADE}, volume 6803 of {\em LNCS}, pages 299--314. Springer, 2011.

\bibitem{Jones72astatistical}
Karen~Spärck Jones.
\newblock A statistical interpretation of term specificity and its application
  in retrieval.
\newblock {\em Journal of Documentation}, 28:11--21, 1972.

\bibitem{holyhammer}
Cezary Kaliszyk and Josef Urban.
\newblock Learning-assisted automated reasoning with {F}lyspeck.
\newblock {\em CoRR}, abs/1211.7012, 2012.

\bibitem{KaliszykU13}
Cezary Kaliszyk and Josef Urban.
\newblock Automated reasoning service for {HOL Light}.
\newblock In Jacques Carette, David Aspinall, Christoph Lange, Petr Sojka, and
  Wolfgang Windsteiger, editors, {\em MKM/Calculemus/DML}, volume 7961 of {\em
  Lecture Notes in Computer Science}, pages 120--135. Springer, 2013.

\bibitem{hhmcs}
Cezary Kaliszyk and Josef Urban.
\newblock {HOL(y)Hammer}: Online {ATP} service for {HOL Light}.
\newblock {\em CoRR}, abs/1309.4962, 2013.

\bibitem{proch}
Cezary Kaliszyk and Josef Urban.
\newblock {PRocH}: Proof reconstruction for {HOL Light}.
\newblock In Maria~Paola Bonacina, editor, {\em CADE}, volume 7898 of {\em
  Lecture Notes in Computer Science}, pages 267--274. Springer, 2013.

\bibitem{EasyChair:74}
Cezary Kaliszyk and Josef Urban.
\newblock Stronger automation for {F}lyspeck by feature weighting and strategy
  evolution.
\newblock In Jasmin~Christian Blanchette and Josef Urban, editors, {\em PxTP
  2013}, volume~14 of {\em EPiC Series}, pages 87--95. EasyChair, 2013.

\bibitem{KuhlweinU12b}
Daniel Kuehlwein and Josef Urban.
\newblock Learning from multiple proofs: First experiments.
\newblock In Pascal Fontaine, Renate~A. Schmidt, and Stephan Schulz, editors,
  {\em PAAR-2012}, volume~21 of {\em EPiC Series}, pages 82--94. EasyChair,
  2013.

\bibitem{KuhlweinBKU13}
Daniel K{\"u}hlwein, Jasmin~Christian Blanchette, Cezary Kaliszyk, and Josef
  Urban.
\newblock {MaSh}: Machine learning for {S}ledgehammer.
\newblock In Sandrine Blazy, Christine Paulin-Mohring, and David Pichardie,
  editors, {\em ITP}, volume 7998 of {\em Lecture Notes in Computer Science},
  pages 35--50. Springer, 2013.

\bibitem{KuhlweinLTUH12}
Daniel K{\"u}hlwein, Twan van Laarhoven, Evgeni Tsivtsivadze, Josef Urban, and
  Tom Heskes.
\newblock Overview and evaluation of premise selection techniques for large
  theory mathematics.
\newblock In Bernhard Gramlich, Dale Miller, and Uli Sattler, editors, {\em
  IJCAR}, volume 7364 of {\em LNCS}, pages 378--392. Springer, 2012.

\bibitem{rehurek10}
Radim {\v R}eh{\r u}{\v r}ek and Petr Sojka.
\newblock Software framework for topic modelling with large corpora.
\newblock In {\em {Proceedings of the LREC 2010 Workshop on New Challenges for
  NLP Frameworks}}, pages 45--50, Valletta, Malta, May 2010. ELRA.

\bibitem{Vampire}
Alexandre Riazanov and Andrei Voronkov.
\newblock The design and implementation of {VAMPIRE}.
\newblock {\em AI Commun.}, 15(2-3):91--110, 2002.

\bibitem{Schapire90}
Robert~E. Schapire.
\newblock The strength of weak learnability.
\newblock {\em Machine Learning}, 5:197--227, 1990.

\bibitem{Sch02-AICOMM}
Stephan Schulz.
\newblock {E - A Brainiac Theorem Prover}.
\newblock {\em AI Commun.}, 15(2-3):111--126, 2002.

\bibitem{KMB}
John Shawe-Taylor and Nello Cristianini.
\newblock {\em Kernel Methods for Pattern Analysis}.
\newblock Cambridge University Press, New York, NY, USA, 2004.

\bibitem{EasyChair:111}
Steffen~Juilf Smolka and Jasmin~Christian Blanchette.
\newblock Robust, semi-intelligible isabelle proofs from atp proofs.
\newblock In Jasmin~Christian Blanchette and Josef Urban, editors, {\em PxTP
  2013}, volume~14 of {\em EPiC Series}, pages 117--132. EasyChair, 2013.

\bibitem{Urban03}
Josef Urban.
\newblock Translating {M}izar for first order theorem provers.
\newblock In {\em MKM}, volume 2594 of {\em LNCS}, pages 203--215. Springer,
  2003.

\bibitem{Urb04-MPTP0}
Josef Urban.
\newblock {MPTP - Motivation, Implementation, First Experiments}.
\newblock {\em Journal of Automated Reasoning}, 33(3-4):319--339, 2004.

\bibitem{Urban06}
Josef Urban.
\newblock {MPTP} 0.2: Design, implementation, and initial experiments.
\newblock {\em J. Autom. Reasoning}, 37(1-2):21--43, 2006.

\bibitem{Urban07}
Josef Urban.
\newblock {MaLARea}: a metasystem for automated reasoning in large theories.
\newblock In Geoff Sutcliffe, Josef Urban, and Stephan Schulz, editors, {\em
  ESARLT}, volume 257 of {\em CEUR Workshop Proceedings}. CEUR-WS.org, 2007.

\bibitem{Urban11-ate}
Josef Urban.
\newblock {An Overview of Methods for Large-Theory Automated Theorem Proving
  (Invited Paper)}.
\newblock In Peter H\"{o}fner, Annabelle McIver, and Georg Struth, editors,
  {\em ATE Workshop}, volume 760 of {\em CEUR Workshop Proceedings}, pages
  3--8. CEUR-WS.org, 2011.

\bibitem{blistr}
Josef Urban.
\newblock {BliStr: The Blind Strategymaker}.
\newblock {\em CoRR}, abs/1301.2683, 2013.

\bibitem{abs-1109-0616}
Josef Urban, Piotr Rudnicki, and Geoff Sutcliffe.
\newblock {ATP} and presentation service for {Mizar} formalizations.
\newblock {\em J. Autom. Reasoning}, 50:229--241, 2013.

\bibitem{US+08}
Josef Urban, Geoff Sutcliffe, Petr Pudl{\'a}k, and Ji\v{r}\'{\i} Vysko\v{c}il.
\newblock {MaLARea SG1 - Machine Learner for Automated Reasoning with Semantic
  Guidance}.
\newblock In Alessandro Armando, Peter Baumgartner, and Gilles Dowek, editors,
  {\em IJCAR}, volume 5195 of {\em LNCS}, pages 441--456. Springer, 2008.

\bibitem{UrbanV13}
Josef Urban and Ji\v{r}\'{\i} Vysko\v{c}il.
\newblock Theorem proving in large formal mathematics as an emerging {AI}
  field.
\newblock In Maria~Paola Bonacina and Mark~E. Stickel, editors, {\em Automated
  Reasoning and Mathematics: Essays in Memory of William McCune}, volume 7788
  of {\em LNAI}, pages 240--257. Springer, 2013.

\end{thebibliography}
\bibliographystyle{plain}

\end{document}